\documentclass[journal]{IEEEtran}

\usepackage{color}
\usepackage{xcolor}
\usepackage{graphicx}
\usepackage{booktabs}
\usepackage{amsmath, amsthm, amsfonts}
\usepackage{mathrsfs}
\usepackage{textcomp}
\usepackage{epstopdf}
\usepackage{multirow}
\usepackage{wrapfig}
\usepackage{subfig}
\usepackage{booktabs} 
\usepackage[ruled]{algorithm2e} 
\usepackage{algorithmicx}  
\usepackage{algpseudocode}
\usepackage{ragged2e}
\usepackage[normalem]{ulem}
\usepackage{cleveref}
\usepackage{bm}
\usepackage[abs]{overpic}

\SetAlFnt{\small}
\SetAlCapFnt{\small}
\SetAlCapNameFnt{\small}
\SetAlCapHSkip{0pt}

\definecolor{green}{rgb}{0, 0.5, 0}
\definecolor{orange}{rgb}{0.6, 0.3, 0.1}
\definecolor{red}{rgb}{1.0, 0.0, 0.0}
\definecolor{teal}{rgb}{0.0, 0.4, 0.4}
\definecolor{purple}{rgb}{0.65,0,0.65}
\definecolor{saffron}{rgb}{0.95,0.75,0.2}
\definecolor{turquoise}{rgb}{0.0,0.5,0.5}
\definecolor{brown}{rgb}{0.5, 0.16, 0.16}
\definecolor{brickred}{rgb}{.6, .2 .1}
\definecolor{coral}{rgb}{1,0.45,0.33}
\definecolor{newcolor}{rgb}{.8,.349,.1}

\begin{document}

\title{Real-Time Spatial Reasoning by Mobile Robots for Reconstruction and Navigation in Dynamic LiDAR Scenes}

\author{Pengdi~Huang, Mingyang~Wang, Huan~Tian, Minglun~Gong, Hao~Zhang, and Hui~Huang$^\dagger$ 

	\thanks{
		This research was funded in parts by GD S\&T Program (2024B01015004), NSFC (U21B2023), ICFCRT (W2441020), Guangdong Basic and Applied Basic Research Foundation (2023B1515120026), DEGP Innovation Team (2022KCXTD025), Shenzhen Science and Technology Program (KJZD20240903100022028, KQTD20210811090044003, RCJC2020071411 4435012), and Scientific Development Funds from Shenzhen University.
	} 
	\thanks{Pengdi Huang, Mingyang Wang, Huan Tian, and Hui Huang are with College of Computer Science and Software Engineering, Shenzhen University, Shenzhen 518060, China (email: alualu628628@gmail.com; michael.gonw@gmail.com; thndy000@gmail.com; hhzhiyan@gmail.com)}
	\thanks{Minglun Gong is with School of Computer Science, University of Guelph, Delft N1G 2W1, Canada (email: minglun@uoguelph.ca)}
	\thanks{Hao Zhang is with School of Computing Science, Simon Fraser University, Burnaby V3J 1A1, Canada (email: haoz@sfu.ca)}  \thanks{$^\dagger$Corresponding author: Hui Huang}
}

\maketitle

\begin{abstract}
Our brain has an inner global positioning system which enables us to sense and navigate 3D spaces in real time. Can mobile robots replicate such a biological feat in a dynamic environment? We introduce the first spatial reasoning framework for real-time surface reconstruction and navigation that is designed for outdoor LiDAR scanning data captured by ground mobile robots and capable of handling moving objects such as pedestrians. Our reconstruction-based approach is well aligned with the critical cellular functions performed by the border vector cells (BVCs) over all layers of the medial entorhinal cortex (MEC) for surface sensing and tracking. To address the challenges arising from blurred boundaries resulting from sparse single-frame LiDAR points and outdated data due to object movements, we integrate real-time single-frame mesh reconstruction, via visibility reasoning, with robot navigation assistance through on-the-fly 3D free space determination. This enables continuous and incremental updates of the scene and free space across multiple frames. Key to our method is the utilization of line-of-sight (LoS) vectors from LiDAR, which enable real-time surface normal estimation, as well as robust and instantaneous per-voxel free space updates. We showcase two practical applications: real-time 3D scene reconstruction and autonomous outdoor robot navigation in real-world conditions. Comprehensive experiments on both synthetic and real scenes highlight our method's superiority in speed and quality over existing real-time LiDAR processing approaches.
\end{abstract} 

\begin{IEEEkeywords}
real-time spatial reasoning, LiDAR, dynamic scenes, outdoor scanning, autonomous navigation
\end{IEEEkeywords}

\IEEEpeerreviewmaketitle


\section{Introduction}
\label{sec:intro}

\IEEEPARstart{H}{umans} and most animals all possess the innate ability to sense and navigate through spatial environments around them in real time. While the complete and precise mechanisms behind such capabilities are not yet fully understood, the Nobel-winning work conducted by John O’Keefe in the 1970s on place cells~\cite{o1971hippocampus,o1976place} has paved the way for understanding the brain’s “inner global positioning system (GPS)”. More recently, several works~\cite{burgess2000predictions,lever2009boundary,hartley2014know,epstein2017cognitive,solstad2008representation} revealed the foundational role of boundary- or border-sensing cells in this GPS, called border vector cells (BVCs) or just border cells. Observations of well-distributed BVCs over all layers of the medial entorhinal cortex (MEC) suggest that recognizing borders and obstacles in a spatial environment may serve as reference frames for place awareness, path finding, and other cellular functions of the spatial representation circuit in the brain~\cite{hartley2000modeling}. These lines of works help explain how the brain forms a cognitive map of the spaces that we visit allowing us to determine our location, find paths between places, and store spatial information for effective navigation.

In this paper, we are intrigued with the question of whether mobile robots can replicate the above biological feats for real-time navigation in dynamic 3D environments, i.e., with the presence of moving objects. The critical cellular function of border sensing and tracking by the brain via BVCs naturally aligns with the geometric surface reconstruction of the navigated dynamic scene. To build a cognitive map for the robot to continually and efficiently explore the environment, moving objects must be properly detected, avoided during navigation, and removed from the reconstructed 3D scene. Accomplishing all these in real time while attaining high reconstruction accuracy poses significant challenges since whatever the scene acquisition mechanism may be, the sensor data obtained is inevitably sparse for complex scenes.

\begin{figure*}[!t]
	\centering
	\hspace*{-2mm}
	\includegraphics[width=\linewidth]{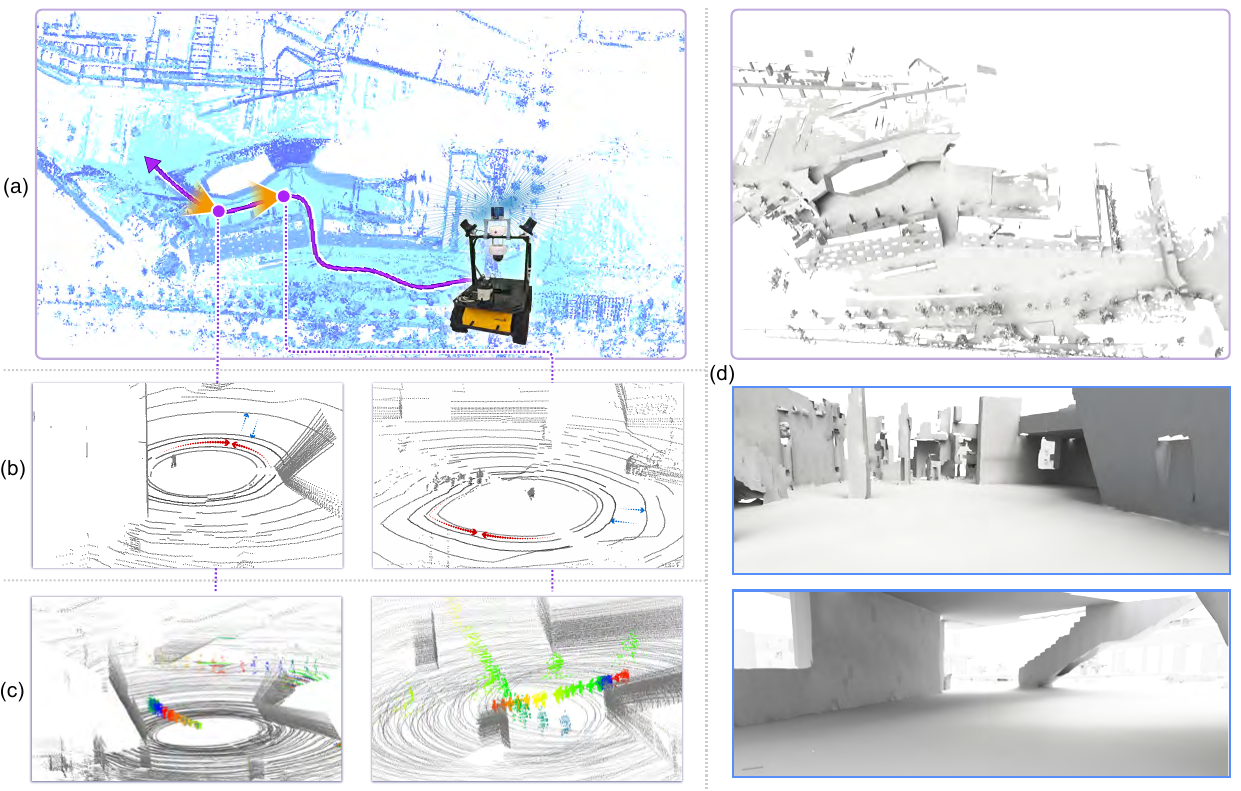}
	\caption{ Real-time robot navigation and reconstruction of large-scale outdoor scenes from LiDAR point clouds, with moving objects such as pedestrians, presents two significant challenges: (b) First challenge: the sparsity and anisotropy (see red directional arrow and blue directional arrow) of LiDAR point cloud data. Due to this, single-frame data cannot completely describe the scene, making it difficult to establish an understanding of unknown objects or obstacles in real time for navigation and reconstruction. (c) Second challenge: proper handling of moving objects. Points captured on moving object at any time are “outdated” quickly for reconstruction, but still need to be tracked as obstacles to be avoided for robot navigation. (d) Our method addresses these challenges by accurately meshing scenes at single-frame and detecting surrounding free spaces in real time. Results showcased here represent the final scene geometry, with moving objects removed. Our system boasts a response speed of 0.1 second, a critical aspect for scene understanding and path navigation by mobile robots, such as the HUSKY, in dynamic environments.}
	\label{fig:teaser}
\end{figure*}

The mobile 3D acquisition mechanism we shall work with is LiDAR (Light Detection and Ranging), which has gained significant popularity in recent years as one of the most widely used 3D spatial sensors for outdoor scene acquisition, especially for navigation~\cite{suger2016terrain,tang2019autonomous,krusi2017driving} and control~\cite{bonatti2020autonomous} by mobile robots~\cite{li2020lidar,pfrunder2017real}. Many robots, especially smaller units, rely only on LiDAR to record spatial maps in the form of point clouds to explore the environment. We focus on LiDAR-only solutions, since they do not require high computing hardware to bear the overhead of reconstructing from image data~\cite{ding2024rescue}. As shown in Fig.~\ref{fig:teaser}, a single frame of online LiDAR data is only a sparse and anisotropic sampling of scene boundaries and an unorganized scan-line-distributed point set. Also, the accumulation of multi-frame point clouds will retain outdated points captured on moving objects. For a LiDAR-only robot to navigate an outdoor scene while avoiding moving obstacles on-the-fly, it is essential to conduct real-time 3D scene reconstruction~\cite{millane2024nvblox} and tracking via a \textit{surface-level} 3D mapping~\cite{pan2022voxfield}, as the BVCs are purported to carry out. 

Besides, A LiDAR scanner can typically acquire a single frame of point cloud scan in $0.1$ second ($10$Hz), generating over $30K$ points (e.g., vlp-16~\cite{park2019curved}) with centimeter accuracy~\cite{bula2020dense}. Interestingly, real-time single-frame surface reconstruction from LiDAR point clouds to accommodate this fast scanning pace would roughly match the typical firing rate of the BVCs~\cite{lever2009boundary,solstad2008representation}. However, to the best of our knowledge, no existing methods have been able to achieve such real-time performance on LiDAR point clouds in converting discrete scan points into continuous spatial coding, i.e, the BVCs role is missing in the general mobile robot scene acquisition and navigation settings. 

\begin{figure*}[!t]
	\centering
	\hspace*{-2mm}
	\includegraphics[width=\linewidth]{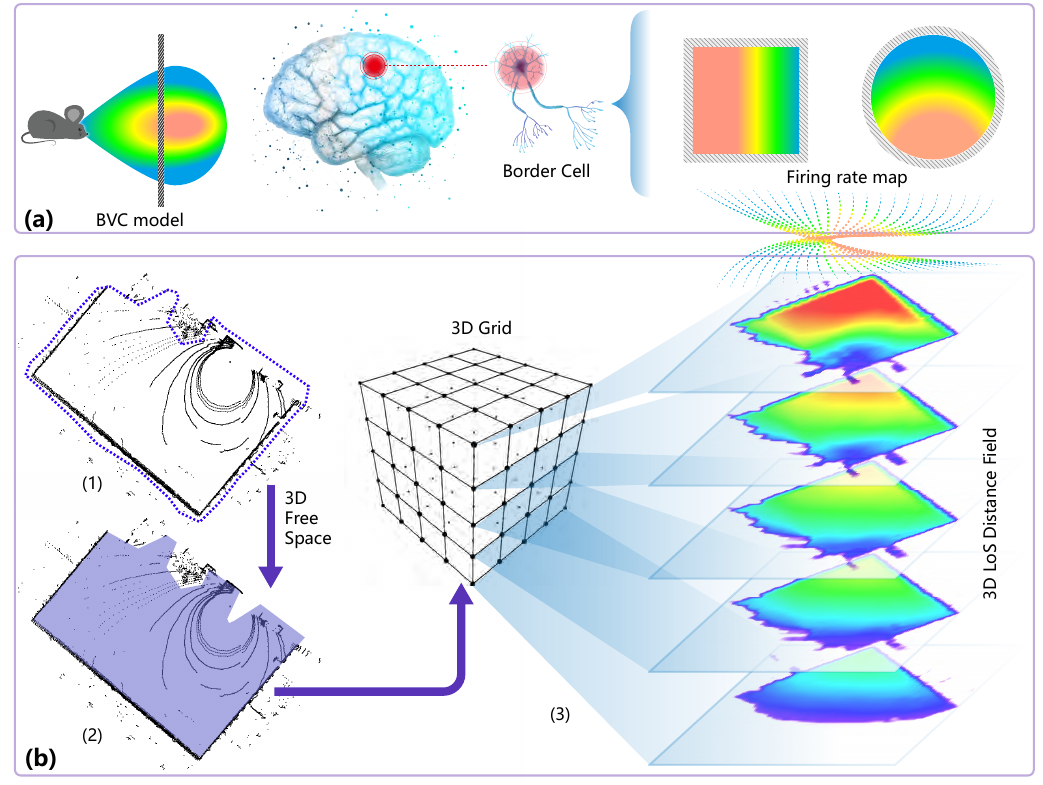}
	\caption{Schematic diagram of our spatial reasoning over online laser scans and resemblance to cellular functions in the brain. Inspired by the BVC model~\cite{lever2009boundary} of border cells in the Hippocampus, we implement the construction of the line-of-sight (LOS) distance field of real-time LiDAR scan point clouds through graphical based pipeline. (a) The foundational role of border-sensing cells in the brain. (b) Replicating the role of BVC in our approach. A single frame of LiDAR scan is shown in (1), whose surface boundary is formed by meshes surrounding the point clouds, as indicated by blue dashed lines. Accordingly, the free space inside the boundary is shown as the blue area in (2), where there are no moving object points, which is important for robot navigation. We employ a 3D grid to encode the 3D free spaces and rely on line-of-sight (LoS) information from the robot to the LiDAR scans, for both real-time surface reconstruction and robust estimation of free spaces, as shown in (3). These operations bear resemblance to spatial sensing and tracking in the entorhinal-hippocampal circuit of the brain, where grid cells receive path planning information about obstacles and boundaries in the scene from border cells~\cite{solstad2008representation}.}
	\label{fig:motivation}
\end{figure*}

Specifically, we present the first method for real-time boundary surface reconstruction and spatial reasoning over online LiDAR scans, designed to handle sparse, anisotropic, and unorganized points acquired from large outdoor scenes with rich geometric variations and moving objects. As shown in Fig.~\ref{fig:motivation}, our spatial reasoning via surface reconstruction must be performed per frame, since a mobile robot needs to identify 3D free space (a.k.a., “safe zone”) while accounting for moving objects, to navigate. At the same time, accumulative free space information across multiple frames enables the removal of moving objects from the final reconstructed 3D scene. To achieve real-time reconstruction, our key insight is to fully leverage data and priors that are readily captured by LiDAR sensors to reason about scene geometry and motion, minimizing unnecessary and potentially costly calculations. To this end, we employ visibility reasoning and utilize line-of-sight (LoS) vectors directly available from LiDAR, to enable real-time normal estimation, surface reconstruction, and robust on-the-fly updates of per-voxel free space identification. Our technical innovations are as follows, each leveraging LoS information in one way or another. 
\begin{itemize}
\item  Observing that all LiDAR points acquired are visible from the robot and mimicking the firing field of BVCs in a single-frame point cloud, we adopt the generalized hidden point removal (GHPR) method~\cite{katz2015visibility}, which was designed for detecting visible points in a point cloud, for single-frame mesh reconstruction and normal estimation. To achieve real-time performance, we exploit the projective nature of GHPR to enable parallel processing. 
\item  To replicate the distance estimation role of BVCs in relation to place cells (see Section~\ref{sec:overview}), we employ the LoS vector for fast weighted normal estimation and discounting of pseudo surfaces that are generated along the LoS directions through parallelization. 
\item Simulating the collaboration between border cells and grid cells over multiple frames, we update per-voxel visibility to identify free spaces. Instead of making binary decisions, which can be error-prone due to acquisition noise and LiDAR data sparsity, we utilize an accumulated LoS distance field to estimate the scene surfaces per frame.
\end{itemize}

\section{Related Work}
\label{sec:rw}

Compared with visual solutions, LiDAR has proven more effective in environments with weak texture or poor illumination~\cite{yang2022robust} and provides better stability under outdoor all-weather conditions~\cite{merriaux2018robust}. As a result, LiDAR-only solutions have been widely adopted by rescue robot~\cite{schwaiger2024ugv}, coal robots~\cite{li2021coal}, field robots~\cite{oliveira2021advances}, industrial inspections~\cite{kaneko2023point}, and autonomous driving~\cite{liu2021efficient}. On the other hand, compared to other depth sensing technologies, such as Microsoft Kinect~\cite{izadi2011kinectfusion}, which is primarily employed in indoor robots, LiDAR poses greater challenges for both scene reconstruction and navigation. 

\subsection{Real-time Scene Reconstruction in Dynamic Scenes}
Real-time scene surface reconstruction from LiDAR scans amounts to simultaneous data acquisition and surface reconstruction, with the latter needing to keep pace with the former. This is necessary, e.g., for obstacle avoidance by a mobile robot~\cite{xu2015autoscanning}. Recent research~\cite{yin2021modeling} suggests that 3D meshes are crucial for robotic interaction with intricate surroundings. To date, on-the-fly acquisition and reconstruction of indoor environments by mobile robots have mostly been achieved with RGBD sensors, e.g., Microsoft Kinect~\cite{zhang2012microsoft,newcombe2011kinectfusion}. However, unlike regularized depth images from RGBD sensors, LiDAR points are captured along laser scanning lines, resulting in gaps between the scan lines that widen as the distance from the sensor increases. 

\begin{figure*}[!t]
	\centering
	\includegraphics[width=\linewidth]{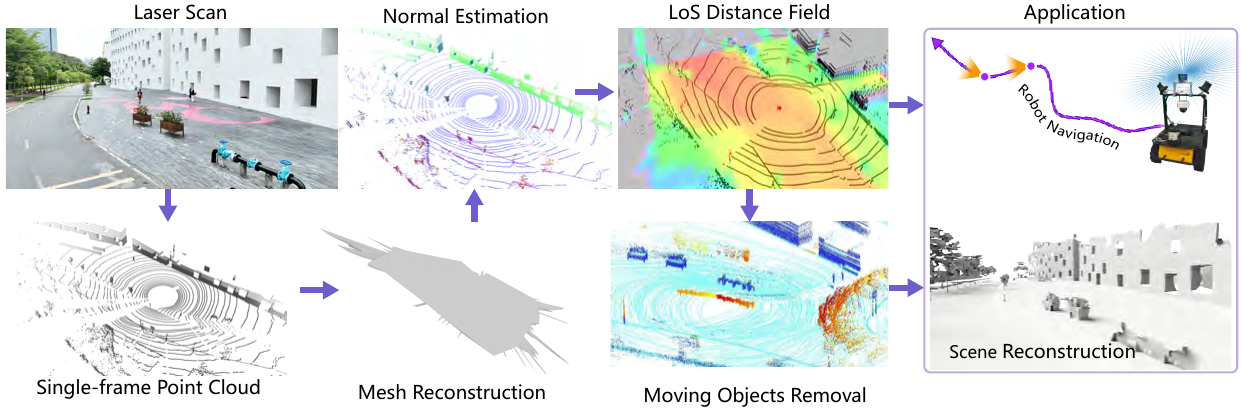}
	\caption{
		An illustration of our real-time spatial reasoning system framework with sample inputs and results.
	}
	\label{fig:overview}
\end{figure*}

Dealing with real-time point clouds in dynamic settings is still at an early stage. 
Classical strategies typically require existing global static point cloud maps to contrast dynamic objects~\cite{krusi2017driving}, which undeniably narrows down practical applications.
Learning-based approaches involving neural fields~\cite{NF_survey} fall far short from providing real-time performance. Existing methods for real-time point cloud reconstruction presume a static scenario.
Behley and Stachniss~\cite{behley2018efficient} propose a surfel-based method for real-time point cloud surface reconstruction. However, this method adopts 2D disk shapes to cover points and thus cannot establish a point-to-point connectivity relationship for polygons.

Recently, Vizzo et al.~\cite{vizzo2021poisson} build a virtual watertight structure for outdoor road scenes and then employ incremental Poisson reconstruction.
However, their method, coined PUMA, is specialized to road scenes and influenced by dynamic objects.
To deal with significant interferences caused by dynamic objects in real-time scanning point clouds, Wu et al.~\cite{wu2024moving} developed M-detector, a straightforward method to remove dynamic objects on-the-fly based on the assumption of occlusion relationship between LiDAR points. They made a connection between the low-latency M-detector and the Magnocellular cells in the lateral geniculate nucleus (LGN) of the human visual system, both operating at a low resolution without border reconstruction. Besides, Wu et al.~\cite{wu2024moving} advocate the use of model-based schemes for moving event detection, attesting to the difficulty of learning-based approaches
both in terms of inference speed and generalizability over out-of-distribution test cases. 

\subsection{Robot Mapping on 3D Signed Distance Fields (SDFs)}
The use of 3D SDF mapping, considered synonymous with surface mesh reconstruction, has become increasingly popular in the field 
of robot perception, e.g., on NVIDIA's Isaac Sim platform~\cite{millane2024nvblox}. 
At the core, the SDF is a numerical representation of a 3D mesh~\cite{pan2022voxfield}, which aids in precise robot navigation~\cite{reijgwart2019voxgraph}.

Even at the grid scale, the accuracy of the SDF~\cite{dang2020graph,han2019fiesta} is significantly higher than that of traditional grid maps that could only represent 2.5D rough terrains~\cite{Fankhauser2018ProbabiliMapping}.
Incremental 3D SDFs, e.g., voxblox~\cite{oleynikova2017voxblox}, use a Euclidean signed distance metric rather than the truncated signed distance metric employed by KinectFusion. 
This allows for calculating signed distances in positions within the scene that are far from the actual surface. However, it is more suitable for dense data captured by RGBD cameras.

For real-time mapping of 3D SDFs and surface reconstruction from laser-scanned point clouds, 
Vizzo et al.~\cite{vizzo2022vdbfusion} follow the technical route of KinectFusion, i.e., using SDF metrics, to estimate the polygonal surface at each single-frame point cloud.
However, real-time reconstruction results by this method, called VDBFusion, are severely restricted by the data quality of the sparse input point clouds.
Zhong et al.~\cite{zhong2023icra} developed a deep network for LiDAR frame reconstruction, coined SHINE-Mapping, which contains an Octree-based feature decoder and a global feature decoder. It outputs an incremental mapping system with predicted grid corner SDF values.
Both of these methods assume static scenes and focus on converting accumulated points into surface models indiscriminately, without special account for moving objects.

To summarize, LiDAR scans acquired in real time do not present shape geometries directly and existing methods are unable to instantly transform the discrete LiDAR points into continuous regions for border or surface reconstruction and 3D mapping. In addition, the stringent speed requirement of real-time surface reconstruction in dynamic scenes, e.g., to allow microsecond-level latency for detecting moving objects, precludes the use of traditional reconstruction schemes for static scenes such as Poisson reconstruction~\cite{kazhdan2006,vizzo2021poisson}, Learning-based approaches~\cite{zhong2023icra}. For this very reason, the M-detector work by Wu et al.~\cite{wu2024moving} advocates the use of model-based, rather than data-driven, schemes for moving event detection, attesting to the difficulty of learning-based approaches both in terms of inference speed and generalizability over out-of-distribution cases. Given these limitations, we argue that LiDAR scan data needs concise and clear scene boundary descriptions to allow real-time surface reconstruction in the single-frame setting. On top of that, we need carefully designed spatial reasoning to merge functional, i.e., navigable, areas in real time for continuous robot navigation.

\section{Overview}
\label{sec:overview}

\paragraph{The mechanisms for spatial representation and mapping}
In formulating our approach, we seek parallels with the mechanisms in the brain for spatial representation and mapping, as illustrated in Fig.~\ref{fig:overview} and detailed below. Neurological studies~\cite{hardcastle2017cell,schmidt2017synaptic,keinath2017environmental} have identified different types of neuronal cells, including place, head direction, grid, and border cells, in the hippocampus and MEC of the brain. They work together to form cognitive maps as spatial memories and their cellular functions were identified by recording the firing of the observed cells while the tested animals were foraging in an open area. 

It was found that place cells fire at specific locations within the area, while head direction cells fire based on head orientation rather than position. Grid cells, on the other hand, fire periodically in a hexagonal grid pattern, and they are activated after place cells and head direction cells~\cite{hartley2014know}. In the context of LiDAR point cloud processing, these functions roughly correspond to simultaneous localization and mapping (SLAM), with a grid-based encoding of the spatial map. Border cells~\cite{lever2009boundary} (BVCs) are widely distributed in the MEC, with their firing fields activated over scene boundaries and parallel to them~\cite{hartley2014know}. BVCs are believed to serve as a reference frame for the other space- and map-sensing cells since fundamentally, these cells would all need to obtain information about obstacles and boundaries in the scene~\cite{solstad2008representation}. Research~\cite{epstein2017cognitive} has also shown that boundaries are strongly correlated with human scene imagination and navigation prediction. 

Inspired by these observations, we aim to replicate the role of BVCs in our approach, focusing on surface (i.e., border) reconstruction to identify scene boundaries, including those of moving obstacles, as shown in Fig.~\ref{fig:motivation}. Furthermore, as BVCs were found to be available before newborns began to explore the world~\cite{donato2023you}, suggesting that their functions do not require training, our approach is entirely model-based, without resorting to data collection or machine learning.

\begin{figure}[!t]
\centering
\includegraphics[width=\linewidth]{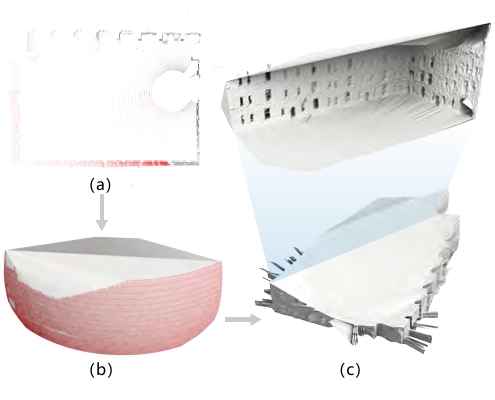}
\caption{
Mesh reconstruction in single-frame mode via GHPR inversion: (a) 3D points captured at the current frame, with a selected region coloured in red; (b) points after GHPR inversion and their convex hull; (c) mesh reconstructed from the convex hull connectivity and partial model corresponding to red point clouds (showing internal reconstruction results).
}
\label{fig:ghpr}
\end{figure}

\begin{figure}[!t]
\centering
\includegraphics[width=\linewidth]{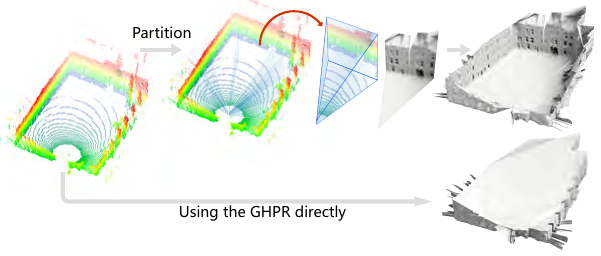}
\caption{Illustration of our proposed partition technique for speeding up single-frame mesh reconstruction. The bottom shows the result of using the GHPR algorithm directly, which produces a global watertight model. The top shows that our method uses radial pie-shaped partitioning to adapt to the radial transformation of GHPR and to avoid the distortion of free space detection. Note that we discarded meshes connected to the viewpoint for displaying the interior surface. In fact, each pie-shaped partition is also reconstructed as a watertight model.}
\label{fig:partition}
\end{figure}

\paragraph{Mesh reconstruction (single-frame)}
In the single-frame mode, we first generate a 3D mesh boundary representation for the 3D points captured at the current frame using the GHPR method introduced by Katz and Tal~\cite{katz2015visibility}; see Fig.~\ref{fig:ghpr}. In our application, we adapt this process to establish point connectivity from the convex hull, which is a non-trivial problem and whose solution allows us to form a collision-free enclosed space based on the visibility from the LiDAR sensor. Furthermore, to achieve {\em real-time\/} performance, we partition the captured points into multiple solid angle sectors from the viewpoint and perform per-sector convex hull construction in parallel; see Fig.~\ref{fig:partition} for an illustration. 

\paragraph{Normal estimation (single-frame)}
Once a mesh is obtained, the next step involves estimating normals at the point samples, thus preserving the boundary structure. We employ a weighted averaging technique based on the surrounding surface normals for this purpose.
This distinctive weighting approach penalizes triangles that are nearly parallel to the line of sight, as these triangles often result in less accurate normals. This approach, in comparison to existing state-of-the-art methods, yields more accurate normal estimation results; see Fig.~\ref{fig:singleframe}. 

\paragraph{Free space reasoning (single-frame and multi-frame)}
Free space is reasoned by discretizing the scene into voxels and calculating the distance between each voxel and the surfaces in the scene along the LiDAR's Line of Sight (LoS). The LoS distance field is computed for each individual frame, and hence takes into account the current locations of moving objects to determine safe spaces.
The distance fields from different frames are then fused to combine observations from various perspectives and to detect moving objects.
Fig.~\ref{fig:movingfree} illustrates the related scenarios and more technical details can be found in Section~\ref{subsec:labeling_scene}.

\paragraph{Application} We also present a real-time surface reconstruction and navigation application of dynamic scenes based on our proposed method, leveraging fast single-frame mesh reconstruction and robot navigation guidance with continuous scene and free space updates across multiple frames. 

Fig.~\ref{fig:overview} provides an overview of the major modes or components of our reconstruction framework. We conduct extensive experiments on both synthetic and real scenes to compare our method to existing approaches applicable to real-time LiDAR reconstruction~\cite{vizzo2021poisson,vizzo2022vdbfusion,zhong2023icra}. We also evaluate our real-time detection and removal of moving objects against state-of-the-art alternatives~\cite{wu2024moving}. The results demonstrate clear advantages of our method in reconstruction quality, while meeting the speed requirements of real-time LiDAR acquisition and navigation. 


\section{Methodology}\label{sec:method}

Our method operates in two primary modes: real-time boundary description in single-frame scan, which reconstructs the 3D scene boundaries for the current field of view, and labeling 3D free space among multiple-frames, which contributes the spatial attributes through multiple perspectives. 

\subsection{Real-time Boundary Description in Single-frame Scan}
\label{subsec:normal_estimation}

Imitating the firing field of boundary cells in the scene, we use a watertight surface model and accurate normal vectors to describe the inner space and boundary of a given single frame point cloud.

\subsubsection{Mesh connectivity for single-frame scan}
\label{subsubsec:mesh_connect}

LiDAR captures 3D points by employing a scanning mechanism that rotates laser beams in a circular or semi-circular pattern. This scanning pattern produces a point cloud with a ring-shaped structure, where the quantity of rings corresponds to the number of laser beams emitted by the LiDAR. Traditional surface reconstruction and normal vector estimation for a query point $p$ require knowledge of geometrically adjacent points, which fails on this ring-shaped structure data. A naive attempt would be to project the single-frame points onto a plane and then connect the triangles. However, this approach results in distorted projections and does not allow for a watertight interior. In this context, we draw inspiration from the GHPR algorithm~\cite{katz2015visibility}. While initially designed for removing occluded points in point clouds, we found that their concept of using a convex hull to establish connectivity among visible points proves effective. It is equivalent to converting the point cloud to a spherical surface and performing spherical meshing, which conforms to the ring-shaped structure of single-frame data. By adapting this idea, we can overcome the difficulties posed by the anisotropic sampling and establish a meaningful connectivity structure for estimating accurate normals in the point cloud.

\begin{figure}[!t]
\centering
\includegraphics[width=\linewidth]{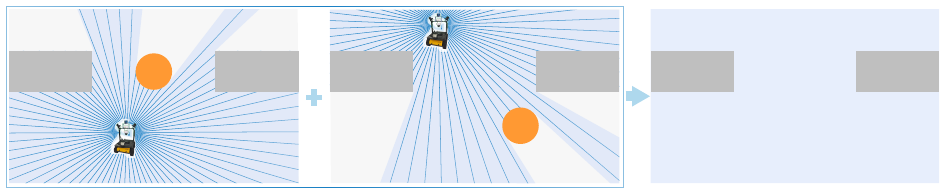}
\caption{
	Illustration of our free space labeling based on visibility reasoning. The round orange object represents a moving object, while the square gray objects depict static surfaces in the scene. Left: the robot measures the LoS distance of the surrounding environment at time $t_1$. Middle: both the robot and the dynamic object move to different locations at $t_2$. The comparison of distance values allows for the detection of the moving object. Right: after consolidating the free space from various viewpoints, the entire scene becomes explored.}
\label{fig:movingfree}
\end{figure}

\begin{figure}[!t]
\centering
\includegraphics[width=\linewidth]{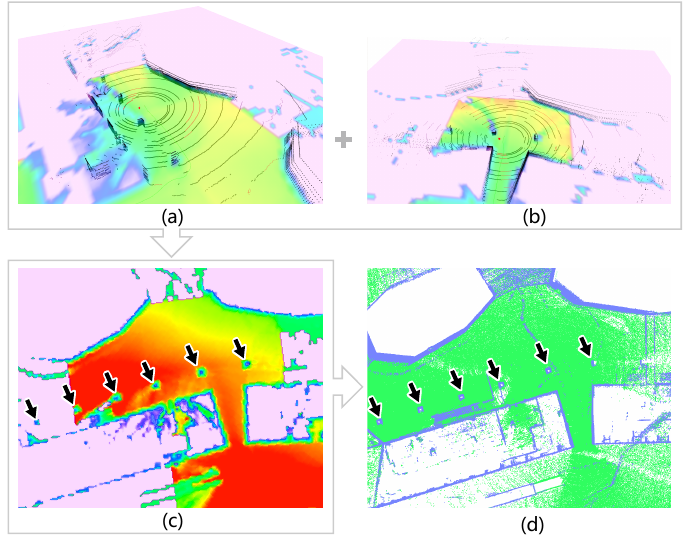}
\caption{
	The LoS distance field of the voxel layer closest to the height of the robot scanner: (a) and (b) the real-time LoS distance fields at two different frames, respectively; (c) the global distance field of this area after a thorough robot scan; note that the values of LoS distances are color-coded in (a)-(c), with pink indicating negative values, blue for values less than half a voxel, and green to red for values ranging from half a voxel to the maximum; (d) the point cloud captured for the same area, color-coded from green to blue based on the elevation value. }
\label{fig:freespaceshow}
\end{figure}

Specifically, here we use $P_t$ to denote the point clouds captured at the $t^{th}$ frame. The coordinates of each point $p \in P_t$ is defined in the local coordinates that are centered at the scanner location $s_t$. $s_t =\left\{{x,y,z}\right\}$ can be obtained from the odometer reading at frame $t$. The GHPR algorithm~\cite{katz2015visibility} works by placing the viewpoint $s_t$ at the center of a sphere and performs radial transformation $F$ on $P_t$ through a kernel function to obtain an inverted point set $P_t'=\{p'\}$:

\begin{equation}
p' = F(p,s_t) = s_t + f_{k}(\|{p - s_t}\| )  \frac{p - s_t}{\|{p - s_t}\|},
\end{equation}
where $p\in P_t$, $f_k(\cdot)$ is the kernel function. In our case, $s_t$ is the origin of the local coordinates for $p$. Hence, the formula is simplified to:
\begin{equation}
p' = f_{k}( \|{p}\| ) \cdot p /\|{p}\|.
\end{equation}
The kernel function $f_k$ usually adopts the mirror mode~\cite{katz2007direct} as follows:
\begin{equation}
f_{k}(\|p\|) = \gamma\mathop{max}\limits_{{q}\in{P}}\ (\|q\|) - \|p\|,
\end{equation}
where $\gamma$ is a scaling factor. Because the scanning range of a single frame point cloud is fixed, we fix $\gamma = 10^{3.7}$ in our experiments. After computing $P'$, we build the convex hull of $P'\cup \{s_t\}$.
Since there is a corresponding relationship between $P$ and $P'$, we can also establish the mesh connectivity of the entire point set $P$ based on the convex hull of $P'\cup \{s_t\}$, as shown in Fig.~\ref{fig:ghpr}. In addition, knowing that all captured points are visible from $s_t$. we point the normals of all triangles in $P$ toward the viewpoint $s_t$.

\subsubsection{Partition for parallel processing and artifacts removal}
\label{subsubsec:partition_parallel}

Performing convex hull calculations directly on the entire point set $P \cup \{s_t\}$  has limitation. Current convex hull algorithms have a non-linear complexity, such as Quick Hull, with a complexity of $O(n\log n)$~\cite{barber1996quickhull}. Consequently, the reconstruction process slows down as the number of points increases. To speed up the process, a simple yet effective approach is used in this paper. It partitions $P$ into $k$ non-overlapping subsets $\{\breve{P}_i\}_{i=1}^k$, where $P = \bigcup_{i=1}^{k} \{\breve{P}_i\}$, and each $\breve{P}_i$ covers a fixed angular range $\theta$ along the horizontal direction. By calculating convex hulls for each of the sets $\breve{P}_i \cup \{s_t\}$ separately, the process is faster and also facilitate parallel processing. As shown in Fig.~\ref{fig:partition}, performing GHPR calculations section by section produces similar 3D representations due to adopting radial based transformation. 

\subsubsection{Normal vector estimation} 
\label{sec:normal}
Before we estimate the normals for all scan points, we first assign a weight to each triangle patch defined by the recovered mesh connectivity.  As we know, 
LiDAR captures 3D points by emitting laser beams and measuring their reflections on object surfaces and the time it takes for the laser beams to travel back. However, when a surface in the scene is parallel to the laser ray, its shape and location cannot be accurately captured by LiDAR. Consequently, any reconstructed surfaces that are close to being coplanar with ray from viewpoint $s_t$ are unreliable and should be discarded. This criterion applies to all non-existent surfaces generated by splitting the point set $P$ into sections, ensuring their removal.
	
Following this assumption, the confidence in the measurement for a triangle is the highest if the triangle is facing the laser scanner and the confidence drops to 0 when the triangle is parallel to laser beams. The confidence weight $w_n$ can then be defined as:
\begin{equation}
w_n = n \cdot (p_{n}-s)/\left\|{p_{n}-s}\right\|.
\end{equation}
The normal vector $n_p$ for each point $p$ is calculated as the weighted average of the triangle normals to which $p$ is associated:
\begin{equation}
n_{p} =  \frac{\sum^{m}_{i=1}{w_{n_i} n_i}}{\|\sum^{m}_{i=1}{w_{n_i} n_i}\|},
\end{equation}
where $m$ is the number of adjacent triangles, while $n_i$ and $w_{n_i}$ denote the normal vector and corresponding weight of triangle $i$.

\begin{figure*}[!t]
	\centering
	\hspace*{-2mm}
	\includegraphics[width=\linewidth]{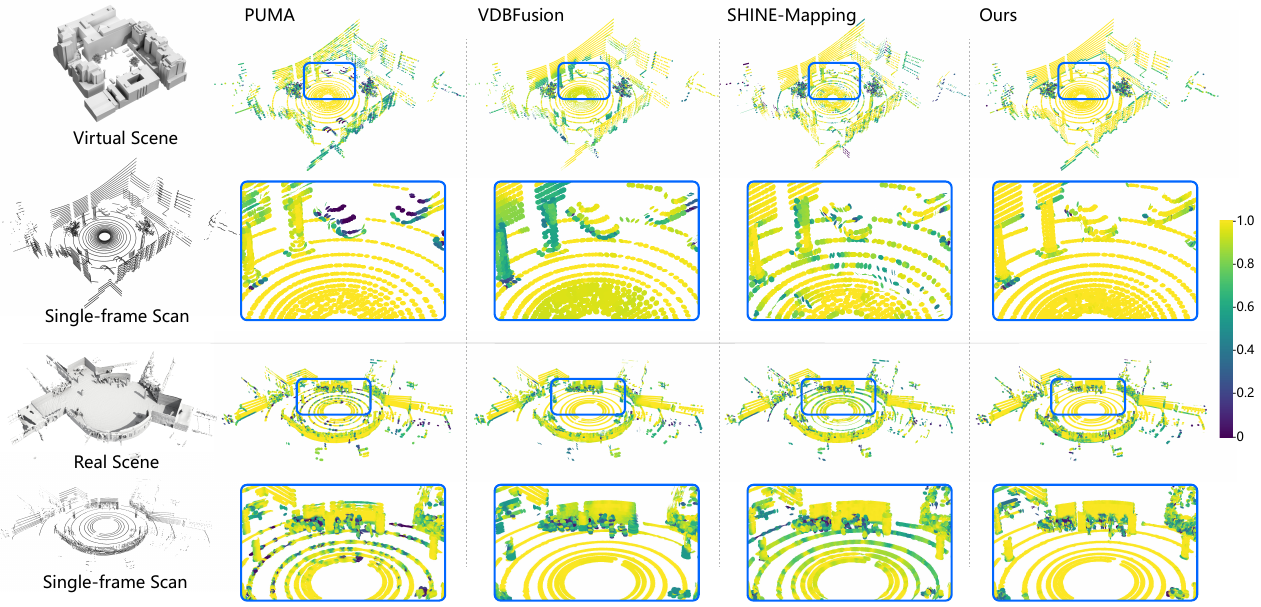}
	\caption{We compare the normal estimation performance of PUMA~\cite{vizzo2021poisson}, VDBFusion~\cite{vizzo2022vdbfusion}, SHINE-Mapping~\cite{zhong2023icra}, and our method on highly non-uniform point clouds captured by a single frame. The top row shows a synthetic scene, while the bottom row shows a real scene. We evaluate the accuracy of the estimated normals by their cosine similarity with the ground truth normals. Higher similarity values are visualized with warmer colors, and lower values with cooler colors. Our method achieves more accurate single-frame point cloud normals.}
	\label{fig:singleframe}
\end{figure*}

Fig.~\ref{fig:singleframe} provides a visual comparison of the estimated normals for each captured point in a specific frame using our approach and other near real-time reconstruction algorithms. The results clearly demonstrate that our approach produces significantly less noisy normals compared to existing methods.

\begin{figure*}[!t]
	\centering
	\includegraphics[width=\linewidth]{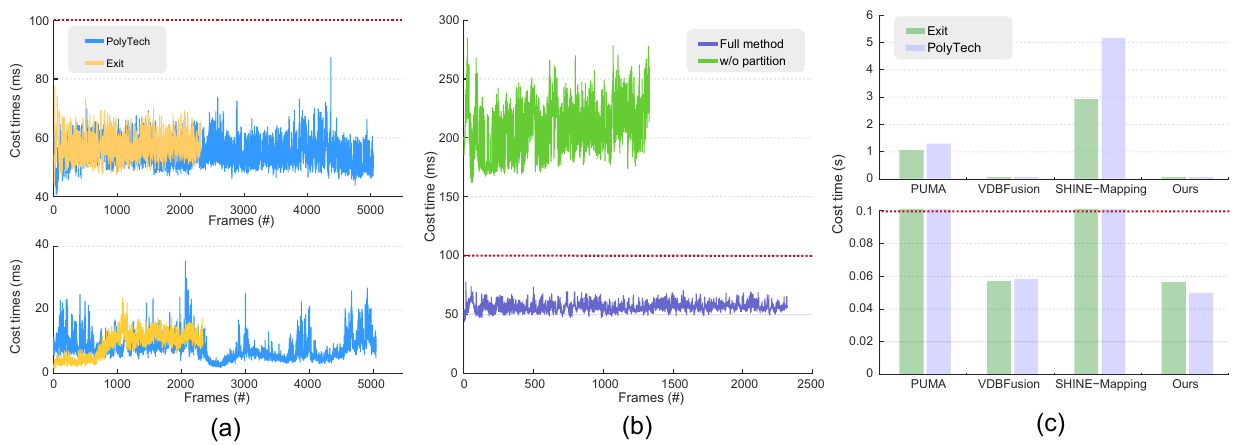}
	\caption{Processing speed evaluation. (a) Time needed of real-time normal estimation (top) and 3D free space labeling (bottom) for processing different frames in two real-world scenes (Exit and PolyTech), respectively. (b) Ablation study on the impact of partition on the Exit scene. (c) The processing time needed by different approaches (SHINE-Mapping~\cite{zhong2023icra}, PUMA~\cite{vizzo2021poisson}, VDBFusion~\cite{vizzo2022vdbfusion}) for single frame processing.}
\label{fig:costtime}
\end{figure*}
\begin{figure}[!t]
	\centering
	\includegraphics[width=\linewidth]{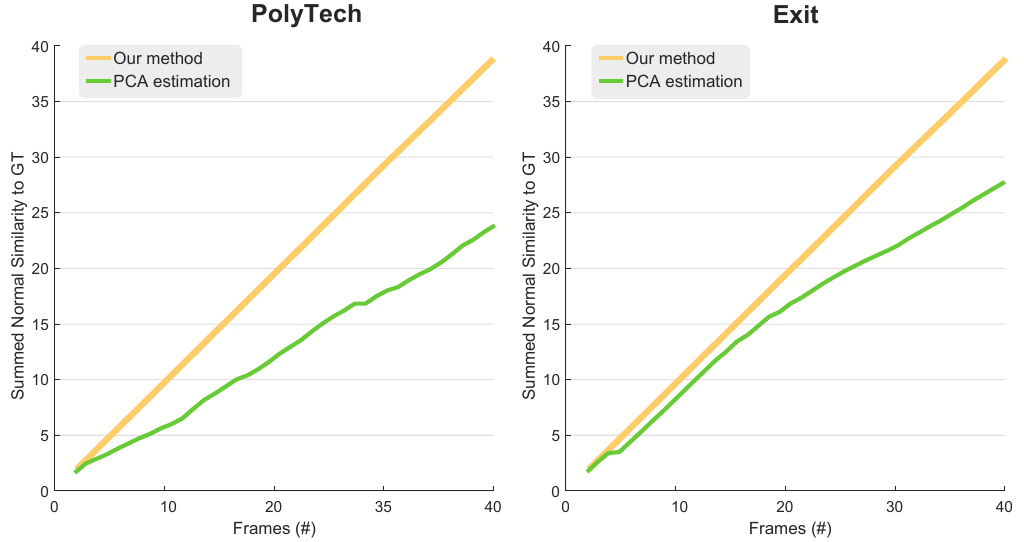}
	\caption{Evaluation results of random 40 frames on real-world PolyTech (left) and Exit (right): the similarity value counts the frame-by-frame accumulation of the cosine value between our normals and the ground truth normals.}
	\label{fig:normalres}
\end{figure}

\subsection{Labeling 3D free space in dynamic scenes}
\label{subsec:labeling_scene}

\emph{3D free space} refers to a set of locations without any objects, meaning that these spaces can be penetrated by the laser beam from a LiDAR scanner. Real-time evaluation of this 3D free space is valuable not only for surface reconstruction but also for robotics, 3D interaction, and virtual reality. Especially in robot navigation, identifying 3D collision-free space establishes a “safe zone” that defines where the robot can pass through without encountering obstacles. Additionally, in dynamic environments, real-time assessment of collision-free safe zones is critical and must be performed at a millisecond pace~\cite{falanga2020dynamic}. Comprehending 3D free space involves extracting the internal space from the entire 3D scene modell~\cite{shalom2010cone}. Similar to the contribution of border cells to grid cells and place cells, the process of delineating this 3D free space comprises two stages: labelling the free space from individual frame data and then consolidating this space across multiple frames.

\subsubsection{Labeling free space from a single frame}

Navigating ground robots through uneven terrain and amidst moving objects presents a formidable challenge. Our approach involves real-time analysis of the captured 3D point cloud data to identify unoccupied space. To achieve this, we discretize the environment into 3D voxels and calculate the distance from each voxel to the nearby object surfaces, as shown in Fig.~\ref{fig:movingfree}. Practically, these distances are computed along the Line of Sight (LoS) of the LiDAR, and therefore, we refer to these distance values as a LoS distance field.
In this field, object surfaces are situated on the 0-value isosurface.
Voxels positioned between the object surfaces and the LiDAR sensor exhibit positive LoS distance values, while those situated behind the object surfaces possess negative values.

To calculate the LoS distance value for a given voxel centered at $q$, we project a ray $R$ from the LiDAR viewpoint $s_t$ toward $q$ and compute the intersection between $R$ and the mesh obtained in the previous section. Assuming the intersection point is $j$, we set the LoS distance $d$ for voxel $q$ at frame $t$ using:
\begin{equation}\label{eq_losdis}
	f_{R}^t(q) =  \|s_t-j\| - \|s_t-q\|,
\end{equation}
\begin{equation}\label{eq_losfield}
	d^t(q) =
	\begin{cases}
		f_{R}^t(q),       & \mbox{if } f_{R}^t(q) \geq 0 \\
		\max(f_{R}^t(q), -\frac{l_{vox}}{2}),    &\mbox{otherwise},
	\end{cases}
\end{equation}
where $l_{vox}$ represents the size of the voxel. The $\max(\cdot,\cdot)$ operation ensures that negative distance values are truncated to a minimum of $-\frac{l_{vox}}{2}$.  This is particularly important because certain voxels might otherwise exhibit high-magnitude negative values, even if they are located in unoccupied space that cannot be detected from viewpoint $s_t$. For a visual representation of the LoS distance field computed from a single frame, please refer to Fig.~\ref{fig:freespaceshow}(a) and~\ref{fig:freespaceshow}(b).

\subsubsection{Fusing LoS distance field across multiple frames}

Because of visibility constraints, a single LiDAR frame only provides information about a portion of the free space within the scene. Therefore, to create an accurate representation of the complete free space throughout the entire scene, it is necessary to combine the LoS distance fields obtained from various frames. In practical terms, we iteratively update the global distance field $D(q)$ using the following approach:
\begin{equation}\label{eq_incremental}
	D^t(q) = \frac{w_{t-1} D^{t-1}(q) + w_{t} d^t(q)} {w_{t-1} + w_{t}}.
\end{equation} 
Here, $w_{t-1}$ and $w_{t}$ represent the weights assigned to the previously fused result $D^{t-1}$ and the new observation $d^t$, respectively. In most cases, setting $w_{t-1}=w_{t}=1$ is sufficient. 

As the mobile LiDAR explores the scene and continuously updates $D^t(q)$, the LoS distance field becomes increasingly refined, where any voxel with $D^t(q)\le\frac{l_{vox}}{2}$ is considered as occupied; see Fig.~\ref{fig:freespaceshow}(c).

\subsubsection{Detection and removal of moving objects}

The generated LoS distance field also facilitates the detection of moving objects. In a scene with only static objects, the values in the LoS distance field increase monotonically as the discovered free space expands while the mobile LiDAR explores the scene from various perspectives. However, when dynamic objects enter previously identified free spaces, the values in the LoS distance field decrease.

\begin{figure*}[!t]
\centering
\includegraphics[width=\linewidth]{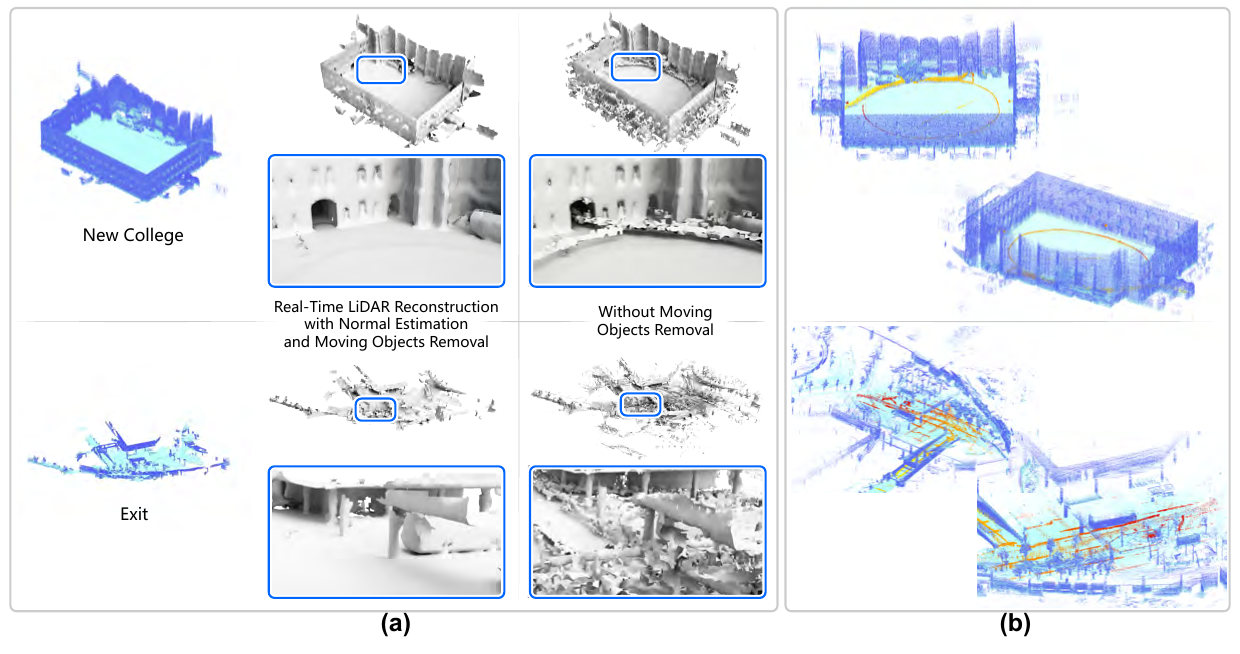}
\caption{
	A gallery showcasing real scenes alongside their real-time reconstruction results. (a) The impact of moving object removal on real-time reconstruction. The raw point cloud containing both moving and static objects are shown on the right, with the corresponding reconstruction results showing spurious shapes. (b) Online per-frame detection and removal of moving objects. Each scene is rendered from two perspectives using all captured points, and static scene points are color-coded from cyan to blue based on elevation values. The detected moving object points are color-coded based on frame IDs.}
\label{fig:dynamicremoval}
\end{figure*}

Hence, comparing $d^t$ and $D^{t-1}$ allows us to identify the point of moving objects.
When the global distance of a query voxel $q$ is $D^{t-1}(q)>\frac{l_{vox}}{2}$, but a measurement $d^t(q)$ is $d^t(q)\le\frac{l_{vox}}{2}$, it indicates that an object is encroaching into a previously free voxel.
Points within this voxel at the time $t$ are labeled as dynamic and are not included in the real-time reconstruction of static scenes.
As shown in Fig.~\ref{fig:dynamicremoval}, our approach can efficaciously remove points from dynamic objects in real time. If not properly excluded, these points can cause significant errors in real-time reconstruction, a phenomenon illustrated in Fig.~\ref{fig:dynamicremoval} and Table~\ref{tab:ablationtable}.

Once points on moving objects are identified and removed, we can fuse all points captured on static surfaces together to reconstruct a high-quality model. In this process, we utilize a voxel-based signed distance function (SDF) framework~\cite{niessner2013real} to perform real-time online modeling for further surface reconstruction application. It is important to highlight that the accurate normals estimated in Section~\ref{sec:normal} play a crucial role in ensuring precise SDF generation.

\section{Implementation}
\label{sec:impl}

\begin{figure}[!h]
\centering
\includegraphics[width=\linewidth]{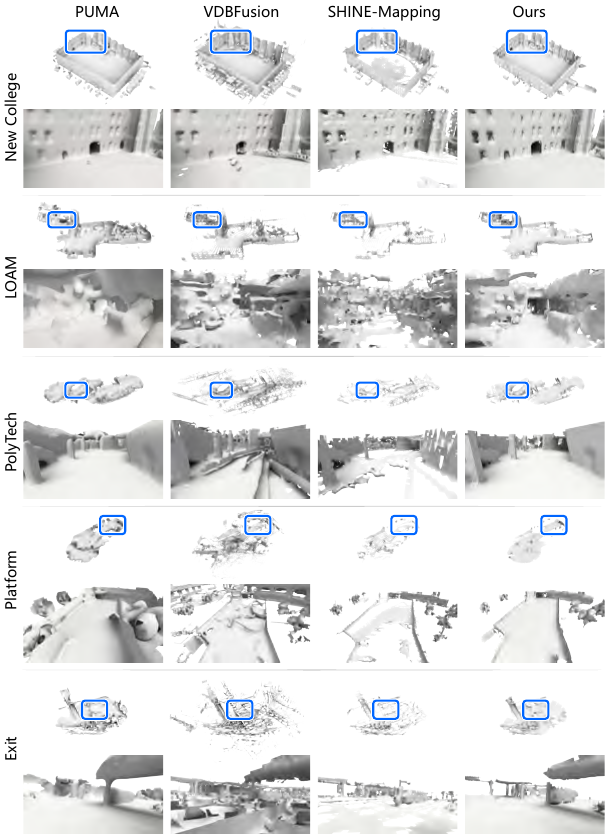}
\caption{
Comparison of multi-frame point cloud reconstruction. Each row represents one of the five real-world scenes, while the columns display the reconstruction results from both state-of-the-art methods and our approach. The blue boxes show areas for the zoomed-in views.}
\label{fig:resmultiframe}
\end{figure}

\begin{figure*}[!htp]
\centering
\includegraphics[width=\linewidth]{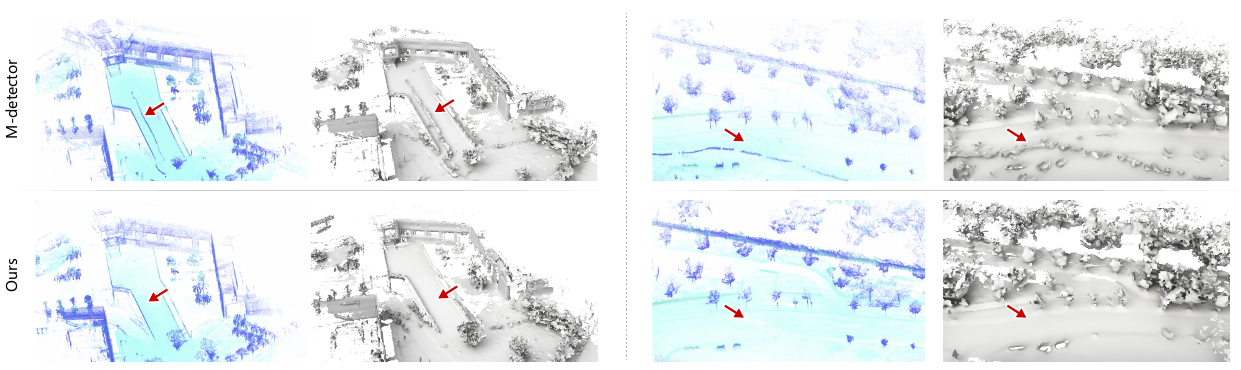}
\caption{
	 Comparison with M-detector~\cite{wu2024moving} on moving object removal. Two examples are shown on the left and the right. For each example, we show the point cloud after moving object removal (color-coded from cyan to blue based on elevation) and the corresponding surface reconstruction result. Our method yields cleaner environments in both examples.}
\label{fig:comparemdetector}
\end{figure*}
\begin{figure*}[!t]
\centering
\includegraphics[width=\linewidth]{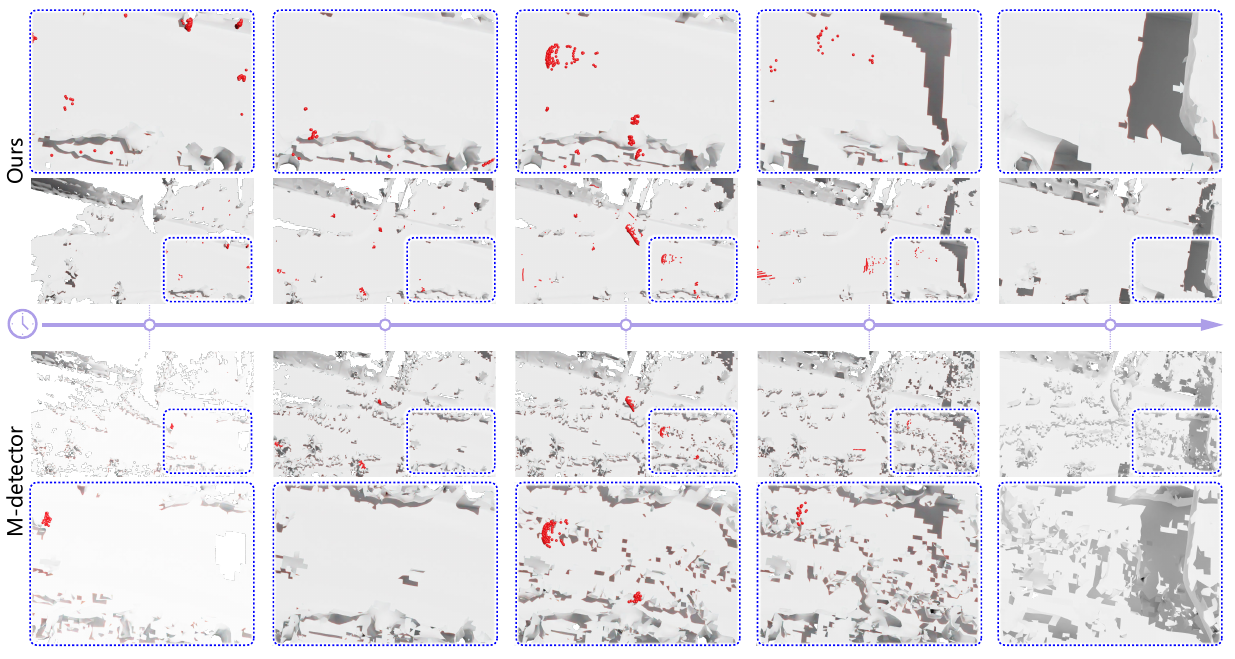}
\caption{
Comparison on real-time reconstruction sequence. The purple arrow in the middle represents the timeline, from which five frames are evenly sampled (marked by white dots). Our real-time reconstruction results (above the timeline) are much cleaner than the results of M-Detector (below the timeline). Red dots indicate the moving object points detected in real-time.}
\label{fig:sequences}
\end{figure*}
\begin{table*}[!ht]
	\caption{Quantitative evaluation of model accuracy from real-time multi-frame reconstruction.}
	\label{tab:multitable}
	\footnotesize
	\centering
	\centering
	\begin{tabular}{c|c|cccccc}
		
		\toprule
		Scene & Method & RMSE(m) $\downarrow$  & Avg. Hausdorff(m) $\downarrow$ & Prec. ($\%$) $\uparrow$ & Recall($\%$) $\uparrow$ & F1($\%$)$\uparrow$ & 95($\%$)Accuracy (m) $\downarrow$\\ 
		
		\midrule
		\multicolumn{1}{c|}{\multirow{4}{*}{New College}}   & \multicolumn{1}{c|}{PUMA} & $0.016546$ & $1.51161$ & $85.8181$ & $80.9002$  & $83.2866$  & $0.004473$ \\
		\multicolumn{1}{c|}{}   & \multicolumn{1}{c|}{VDBFusion} & $0.0192$ & $1.58905$ & $89.174$ & $77.1687$  & $82.7382$  & $0.004661$ \\
		\multicolumn{1}{c|}{}   & \multicolumn{1}{c|}{SHINE-Mapping} & $0.021157$ & $1.15611$ & $93.383$ & $68.8238$  & $79.2441$  & $0.00362$ \\
		\multicolumn{1}{c|}{}   & \multicolumn{1}{c|}{Ours} & $\mathbf{0.013485}$ & $\mathbf{0.91688}$ & $\mathbf{96.9324}$ & $\mathbf{84.1233}$  & $\mathbf{90.0747}$  & $\mathbf{0.000594}$ \\

		\midrule		
		\multicolumn{1}{c|}{\multirow{4}{*}{LOAM}} & \multicolumn{1}{c|}{PUMA} & $0.093625$	& $3.64489$ & $38.4148$ & $60.846$ & $47.0959$ & $0.158062
		$ \\ 
		\multicolumn{1}{c|}{}   & \multicolumn{1}{c|}{VDBFusion} & $0.090907$	& $	15.945$	& $	52.8647$	& $	48.6517	$	& $50.6708$	& $	0.12095
		$ \\
		\multicolumn{1}{c|}{}   & \multicolumn{1}{c|}{SHINE-Mapping} & $0.077685$	& $	2.74752	$	& $62.2999$	& $	57.0558	$	& $59.5626$	& $	0.117345$ \\ 
		\multicolumn{1}{c|}{}   & \multicolumn{1}{c|}{Ours} & $\mathbf{0.055349}$ & $\mathbf{	0.847758}$ & $\mathbf{	77.9534	}$ & $\mathbf{63.6478}$ & $\mathbf{	70.078}$ & $\mathbf{0.017305}$ \\

		\midrule		
		\multicolumn{1}{c|}{\multirow{4}{*}{PolyTech}} & \multicolumn{1}{c|}{PUMA} & $0.021203$ & $	5.66566$ & $58.7237$ & $\mathbf{57.0534}	$ & $57.8765$ & $	0.169302$ \\ 
		\multicolumn{1}{c|}{}   & \multicolumn{1}{c|}{VDBFusion} & $0.026093$ & $	35.8998	$ & $48.6477$ & $	30.8013$ & $	37.7201$ & $	8.0583
		$ \\
		\multicolumn{1}{c|}{}   & \multicolumn{1}{c|}{SHINE-Mapping} &$ 0.040874$ & $	5.3761$ & $	73.452	$ & $48.2209$ & $	58.2204	$ & $0.145258
		$ \\ 
		\multicolumn{1}{c|}{}   & \multicolumn{1}{c|}{Ours} & $\mathbf{0.012675}$ & $\mathbf{	1.58901}$ & $\mathbf{	93.7098}$ & $56.509$ & $\mathbf{	70.5031}$ & $\mathbf{	0.001979}	$ \\

		\midrule		
		\multicolumn{1}{c|}{\multirow{4}{*}{Platform}} & \multicolumn{1}{c|}{PUMA} & $0.019576$ & $	3.5933$ & $	61.6441	$ & $69.7413$ & $	65.4432	$ & $0.136851	$ \\ 
		\multicolumn{1}{c|}{}   & \multicolumn{1}{c|}{VDBFusion} & $0.024804$ & $		39.7905	$ & $	45.3218	$ & $	37.1458	$ & $	40.8285	$ & $	21.3448$ \\
		\multicolumn{1}{c|}{}   & \multicolumn{1}{c|}{SHINE-Mapping} & $0.030681$ & $	6.16588$ & $	77.4801	$ & $57.1554$ & $	65.7836	$ & $0.114543$ \\ 
		\multicolumn{1}{c|}{}   & \multicolumn{1}{c|}{Ours} & $\mathbf{0.012499}$ & $\mathbf{	0.781985}$ & $\mathbf{	96.0936}$ & $\mathbf{	72.9949}$ & $\mathbf{	82.9665}$ & $\mathbf{	0.000821}
		$ \\ 
		
		\midrule		
		\multicolumn{1}{c|}{\multirow{4}{*}{Exit}} & \multicolumn{1}{c|}{PUMA} & $0.024647$ & $4.94264$ & $61.2038$ & $77.0675$  & $68.2256$  & $0.062086$ \\ 
		\multicolumn{1}{c|}{}   & \multicolumn{1}{c|}{VDBFusion} & $0.02829$ & $36.7906$ & $43.2164$ & $55.0413$  & $48.4173$  & $15.1074$ \\
		\multicolumn{1}{c|}{}   & \multicolumn{1}{c|}{SHINE-Mapping} & $0.036915$ & $8.11109$ & $70.5317$ & $75.2236$  & $72.8022$  & $0.061108$ \\ 
		\multicolumn{1}{c|}{}   & \multicolumn{1}{c|}{Ours} & $\mathbf{0.015323}$ & $\mathbf{1.32439}$ & $\mathbf{92.1178}$ & $\mathbf{85.4479}$ & $\mathbf{88.6576}$  & $\mathbf{0.003038}$ \\ 
		
		
		\bottomrule
	\end{tabular}%
\end{table*}

\begin{table*}[!ht]
	\caption{Comparative evaluation of model accuracy on the impact of moving object detection and removal.}
	\label{tab:ablationtable}
	\footnotesize
	\centering
	\renewcommand{\arraystretch}{1.2}
	\centering
	\begin{tabular}{c|c|cccccc}
		
		\toprule
		Scene & Method & RMSE(m) $\downarrow$  & Avg. Hausdorff(m) $\downarrow$ & Prec. ($\%$) $\uparrow$ & Recall($\%$) $\uparrow$ & F1($\%$)$\uparrow$ & 95($\%$)Accuracy (m) $\downarrow$\\ 
		
		\midrule
		\multicolumn{1}{c|}{\multirow{2}{*}{New College}}   & \multicolumn{1}{c|}{w/o moving object removal } & $0.015272$ & $1.82866$ & $91.4738$ & $79.6639$  & $85.1613$  & $0.0019$ \\ 
		\multicolumn{1}{c|}{}   & \multicolumn{1}{c|}{Using M-detector} & $0.015423$ & $1.06513$ & $95.6089$ & $81.9296$  & $88.2423$  & $0.001449$ \\
		\multicolumn{1}{c|}{}                            & \multicolumn{1}{c|}{Our full method} & $\mathbf{0.013485}$ & $\mathbf{0.91688}$ & $\mathbf{96.9324}$ & $\mathbf{84.1233}$  & $\mathbf{90.0747}$  & $\mathbf{0.000594}$ \\

		\midrule
		\multicolumn{1}{c|}{\multirow{2}{*}{LOAM}}   & \multicolumn{1}{c|}{w/o moving object removal} & $0.065148$  & $\mathbf{0.577518}$  & $73.5284$  & $58.1333$  & $	64.9308$  & $0.01796$ \\ 
		\multicolumn{1}{c|}{}   & \multicolumn{1}{c|}{Using M-detector} & $0.236745$ & $2.53495$ & $32.2705$ & $35.7053$  & $33.9011$  & $0.225640$  \\ 
		\multicolumn{1}{c|}{}                            & \multicolumn{1}{c|}{Our full method} & $\mathbf{0.055349}$ & $0.847758$ & $\mathbf{77.9534}$ & $\mathbf{	63.6478}$ & $\mathbf{70.078}$ & $\mathbf{0.017305}$ \\ 		

		\midrule
		\multicolumn{1}{c|}{\multirow{2}{*}{PolyTech}}   & \multicolumn{1}{c|}{w/o moving object removal} & $0.016195$  & $	4.83835	$  & $84.02	$  & $47.2916$  & $	60.5193$  & $	0.007535$ \\
		\multicolumn{1}{c|}{}   & \multicolumn{1}{c|}{Using M-detector} &$0.014480$ & $5.38009$ & $90.9631$ & $\mathbf{57.6348}$  & $\mathbf{70.5614}$  & $0.0034547$  \\ 
		\multicolumn{1}{c|}{}                            & \multicolumn{1}{c|}{Our full method} & $\mathbf{0.012675}$ & $\mathbf{1.58901}$ & $\mathbf{93.7098}$ & $56.509$ & $70.5031$ & $\mathbf{0.001979}	$  \\ 	
		
		\midrule
		\multicolumn{1}{c|}{\multirow{2}{*}{Platform}}   & \multicolumn{1}{c|}{w/o moving object removal} & $0.015402$  & $	2.27641$  & $	91.2964	$  & $64.7412	$  & $75.7591$  & $	0.002322$ \\ 
		\multicolumn{1}{c|}{}   & \multicolumn{1}{c|}{Using M-detector} & $0.013527$ & $2.58599$ & $95.5155$ & $\mathbf{76.585}$  & $\mathbf{85.0091}$  & $0.001197$  \\ 
		\multicolumn{1}{c|}{}                            & \multicolumn{1}{c|}{Our full method} & $\mathbf{0.012499}$ & $\mathbf{0.781985}$ & $\mathbf{96.0936}$ & $72.9949$ & $82.9665$ & $\mathbf{0.000821}$  \\ 	
		
		\midrule
		\multicolumn{1}{c|}{\multirow{2}{*}{Exit}}   & \multicolumn{1}{c|}{w/o moving object removal} & $0.018751$ & $5.75306$ & $79.1931$ & $74.1899$  & $76.6099$  & $0.018163$ \\ 
		\multicolumn{1}{c|}{}   & \multicolumn{1}{c|}{Using M-detector} & $0.017236$ & $7.27246$ & $89.4401$ & $80.3193$  & $84.6347$  & $0.005928$  \\ 
		\multicolumn{1}{c|}{}                            & \multicolumn{1}{c|}{Our full method} & $\mathbf{0.015323}$ & $\mathbf{1.32439}$ & $\mathbf{92.1178}$ & $\mathbf{85.4479}$ & $\mathbf{88.6576}$  & $\mathbf{0.003038}$ \\ 		
		\bottomrule
	\end{tabular}%
\end{table*}

\begin{figure*}[!htp]
\centering
\includegraphics[width=\linewidth]{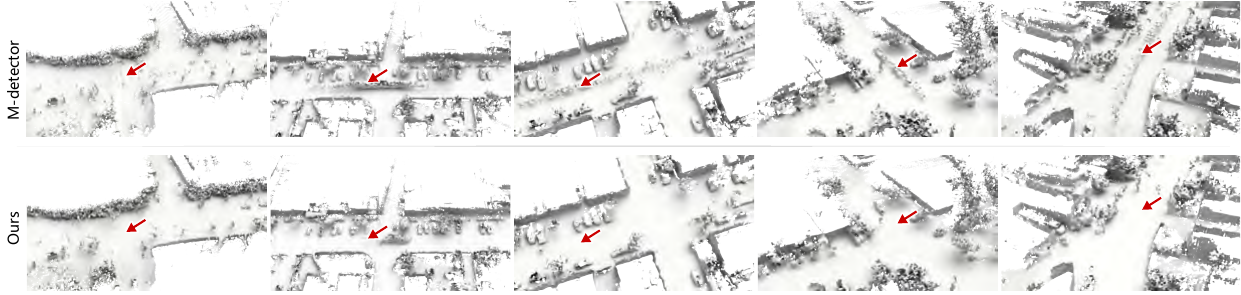}
\caption{
	 A gallery showcasing reconstruction quality comparison with M-detector on the KITTI dataset. Our method (bottom) handles crowded traffic scenes much better than the M-detector (top), leading to more accurate scene structures of surrounding environments.}
\label{fig:publicgallery}
\end{figure*}

\begin{figure*}[!t]
\centering
\includegraphics[width=\linewidth]{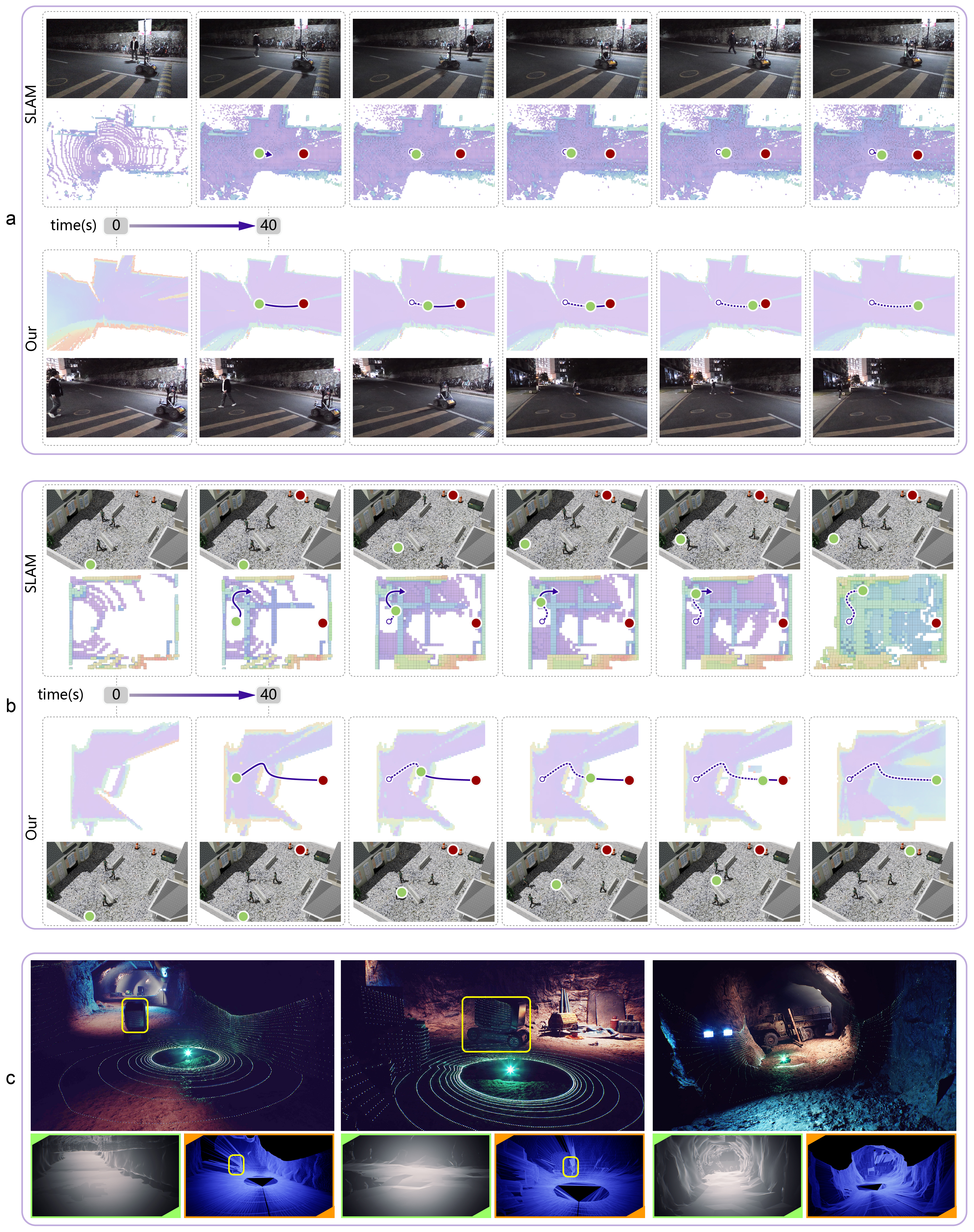}
\caption{
 Robot navigation applications. (a) and (b) Robot navigation comparison experiment in real environment and in synthetic environment, respectively. (c) Scene exploration in a synthetic working coal tunnel.}
\label{fig:robotnavigation}
\end{figure*}

\paragraph{System setup}
The input of our method is the real-time output of LiDAR-only SLAM, e.g., the current trajectory point and its corresponding single-frame points. Theoretically, various SLAM methods, including LOAM~\cite{zhang2014loam}, LIO-SAM~\cite{shan2020lio}, and Lego-LOAM~\cite{shan2018lego}, can be integrated into our method. Our method must accommodate the errors generated by various SLAMs. For evaluation purposes, we used KISS-ICP~\cite{vizzo2023kiss} as an alternative to SLAM to estimate the frame poses for all methods, which minimized the influence of different SLAM algorithms on the reconstruction results. For robot navigation applications, we implemented our proposed reconstruction method on the HUSKY robot platform, which is equipped with a Hesai Pandar LiDAR scanner. The LiDAR setup consisted of 16 laser lines for SLAM processing.

\paragraph{Parameter setting}
In our experiments, we implemented the ray intersection operation based on Embree~\cite{wald2014embree}. We set the parameters $w_{t-1}$ and $w_t$ in Eq.\ref{eq_incremental} to $1$, and the voxel size $l_vox$ in Eq.\ref{eq_losfield} to $0.5m$. Besides, the fixed angle used in partitioning is set $pi⁄6$, and the surface update radius of the robot for real-time reconstruction is set to $30.0m$.

\paragraph{Dataset and metric}
We evaluated our proposed method using both synthetic and real datasets. The real datasets consist of five scenes: New College, LOAM, PolyTech, Platform, and Exit. New College and LOAM are widely used public datasets and serve as common benchmarks for assessing and comparing different algorithms and methods. New College is a hand carried LiDAR dataset~\cite{ramezani2020newer} and LOAM is a robot mobile laser point cloud dataset generated by VLP-16 LiDAR~\cite{zhang2014loam}. The last three datasets are campus scenes that we collected; see Fig.~\ref{fig:freespaceshow}, we conducted testing on the large public outdoor dataset KITTI~\cite{geiger2012we}. This dataset consists of real-time LiDAR scanning point cloud data captured in an urban traffic environment.

The synthetic datasets also consist of four distinct scenes, namely Ground, Center, Square, Coal tunnel, each featuring different scenarios with varying off-ground objects and density distributions. As depicted in Fig.~\ref{fig:robotgallery}, the Ground scene in ROS Gazebo is an open-source synthetic environment provided by the HUSKY platform, characterized by a sparsely populated setting. The Center and Community scenes in ROS Gazebo represent scenarios with moderately populated buildings, while the Coal tunnel in UE5 platform reflect complex environments with moving objects.

The normal vector is used to reveal the inside and outside of the scene boundary and represents the local mesh construction performance. To quantitatively assess the accuracy of normals generated using single-frame mesh reconstruction, we employ the dot product between the estimated and ground truth normal vectors as an evaluation metric. The ground truth model is densely sampled, ensuring that a nearby ground truth point can be identified for each captured point. For the evaluation of 3D models obtained through multi-frame point cloud reconstruction, we utilize a range of evaluation metrics, including point-to-point RMSE (Root Mean Square Error), average Hausdorff distance, F1-score, and accuracy. These metrics provide a comprehensive assessment of the reconstructed 3D model’s quality and accuracy. For each real scene, its ground truth model is acquired through the registration of all point clouds, followed by manual removal of points corresponding to moving objects.

\section{Experimental Results}
\label{sec:results}

\subsection{Processing Time}
Fig.~\ref{fig:costtime}(a) plots the processing time of real-time boundary description (Section~\ref{subsec:normal_estimation}) and labeling 3D free space (Section~\ref{subsec:labeling_scene}) across different frames for two real-world scenes: Exit and PolyTech. In our real-time system, these processes are implemented as robot operating system (ROS) nodes and run in parallel. The results demonstrate our approach consistently achieves approximately 10fps for both scenes, ensuring real-time processing without data loss. Additionally, the figure includes an ablation study which reveals that the processing time increases at least fourfold without partitioning the captured points into pie-shaped sections. This highlights the efficacy of our partitioning strategy in accelerating the process.

Fig.~\ref{fig:costtime}(c) presents a comparison of the computational time required by different approaches for single-frame point cloud normal estimation. Notably, our method demonstrates efficient processing, achieving 10 frames per second (fps), matching the upper limit of the laser radar emission frequency (10Hz).
Considering that unmanned vehicles can reach speeds of up to 20$m/s$, such processing frequency is important for ensuring timely and safe responses in autonomous driving applications. In this context, only our method and TSDF-based approaches, such as VDBFusion, can meet the demanding real-time processing requirements.

\subsection{Single-frame Normal Estimation Accuracy}
To evaluate the online boundary description ability, we adopt the normal vector as the geometric evaluation target. Fig.~\ref{fig:singleframe} illustrates the evaluation of single-frame point cloud normal estimation accuracy, using both synthetic and real scans. The synthetic scan consists of 34,246 points, while the real scan was captured using a Heisai 40P LiDAR and contains 27,304 points. The ground truth of real scan is the normal vector of the offline NDC~\cite{chen2022neural} generation model. To enhance visualization, we performed voxel downsampling on the estimated point cloud normals, using a voxel size of 0.5m. This downsampling resulted in a reduced point count of 4,428.

From the figure, it is evident that our method produces normal vectors that closely match the ground truth. Notably, VDBFusion~\cite{vizzo2022vdbfusion}, an improved TSDF method derived from KinectFusion~\cite{newcombe2011kinectfusion}, is also included for comparison. This comparison highlights the limitations of the KinectFusion series in single-frame point normal estimation.

Our method excels in local geometry analysis without the need to assume a specific neighborhood size. This adaptability allows for accurate normal vector estimation on non-uniformly sampled scan lines, effectively addressing the challenges presented by anisotropic and sparse single-frame LiDAR point clouds. Additionally, our single-frame normal estimation approach eliminates the need to specify a voxel size or observation scale, making it adaptable to various devices and scenes.

Fig.~\ref{fig:normalres} shows an accuracy evaluation of normal estimation on real-world PolyTech and Exit scenes. 
We conducted a random sampling of 40 frames from the scan sequences of these scenes. Our method consistently outperforms PCA, which relies on selecting the 10 nearest points as the local neighborhood. Our similarity accumulation plot exhibits faster growth rates compared to those from the baselines, and the growth rate reaches an average of 0.9565 and 0.9613 per frame, while the growth rate of PCA estimation baseline only reaches an average of 0.6684 and 0.5954 per frame.

\begin{figure*}[!t]
\centering
\includegraphics[width=\linewidth]{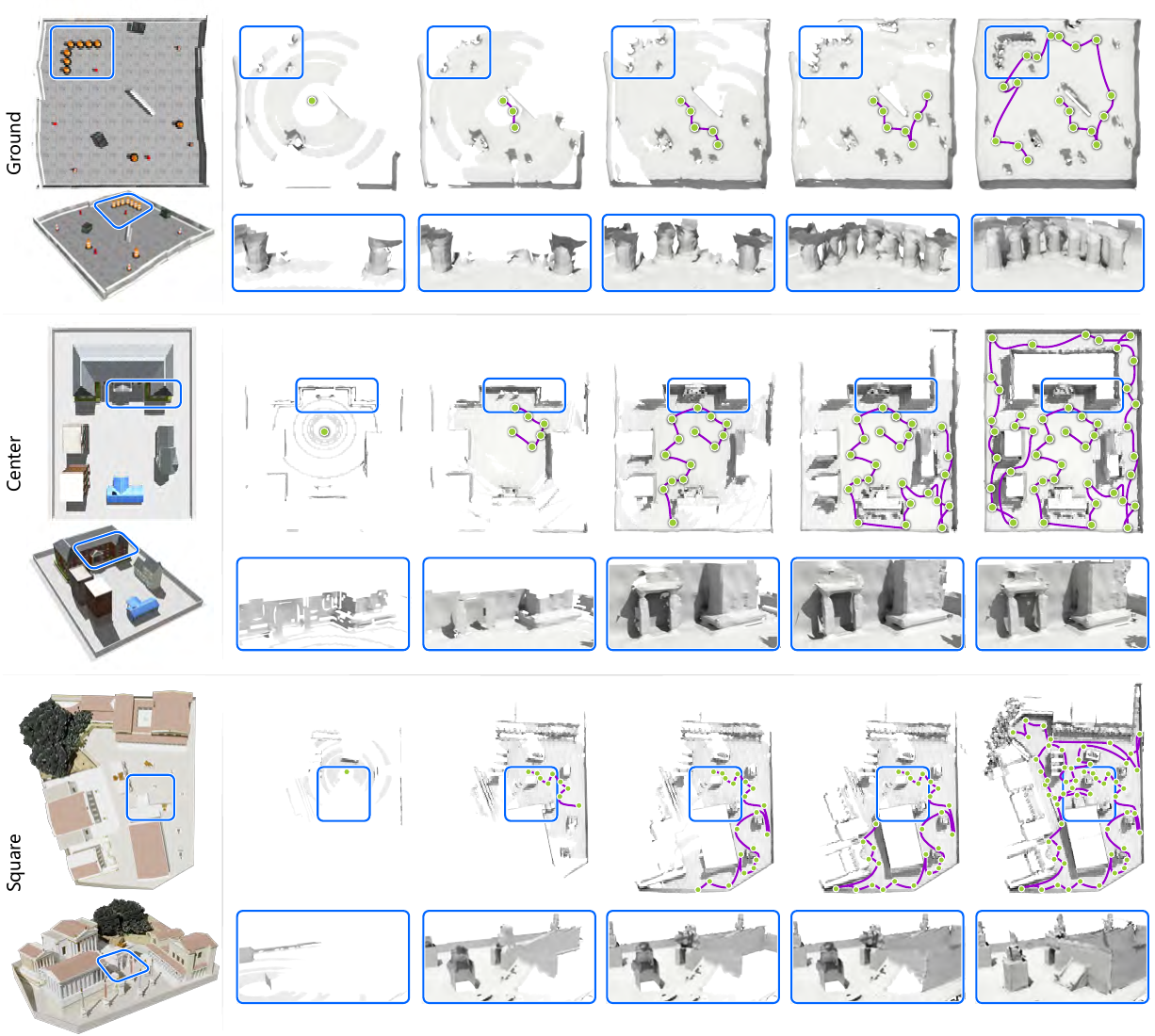}
\caption{
A gallery showcasing our system's application on real-time navigation and construction. The first column displays synthetic scenes, while columns two to five illustrate the process of autonomous data acquisition and real-time surface reconstruction. The blue boxes provide zoomed-in views of the reconstructed modeling. The purple curve traces the autonomous navigation path, while green dots represent the real-time planning target points for robot navigation.}
\label{fig:robotgallery}
\end{figure*}

\subsection{Application on real-time scene surface reconstruction}
At the foundation of our reconstruction paradigm is the classical marching cube based real-time reconstruction pipeline by Niessner et al.~\cite{niessner2013real}, which utilizes estimated normal and voxel SDF value. The SDF value of a voxel is computed from the mean normal of the non-moving object points within the voxel, which is essentially the transformation of the SDF value through the mean plane.
Table~\ref{tab:multitable} presents a comparison of our multi-frame point cloud real-time reconstruction method with PUMA~\cite{vizzo2021poisson}, VDBFusion~\cite{vizzo2022vdbfusion}, and SHINE-Mapping~\cite{zhong2023icra} across five real-world scenes. Our models exhibit the smallest errors in terms of RMSE, average Hausdorff distance, and accuracy. Additionally, our method achieves larger F-scores, indicating superior performance in terms of precision and recall. These results highlight the suitability of our method for real-time outdoor reconstruction. The improved reconstruction quality can be attributed to two key factors. First, our single-frame calculations accurately estimate the surface normals, leading to more accurate reconstruction. Second, our method effectively removes points captured from dynamic objects, further enhancing the quality of the generated model.

Fig.~\ref{fig:resmultiframe} provides a visual comparison among results generated by different approaches. Our method produces clearer and more precise models. Importantly, it also demonstrates robustness against real-time noise, which is crucial for reconstruction tasks and robot perception.

Table~\ref{tab:ablationtable} and Fig.~\ref{fig:dynamicremoval} present an ablation study on the impact of moving object detection and removal in real-time multi-frame point cloud reconstruction. The results demonstrate that moving objects introduce significant noise that degrades reconstruction quality. By effectively eliminating points associated with moving objects, our method exhibits a notable improvement in the quality of the reconstructed surface model compared to the version without such detection.

To assess the proposed moving object detection strategy with existing approaches, such as M-detector~\cite{wu2024moving}, Fig.~\ref{fig:comparemdetector} and Table~\ref{tab:ablationtable} compare our full method with a version that adopts M-detector. The M-detector method relies on an intricate occlusion relationship between the scene foreground and background. Its performance is often limited by the large scale of outdoor environments and the pose estimation error by SLAM. In contrast, our method is more robust to noise and the dynamic structure of outdoor scenes.

Fig.~\ref{fig:sequences} illustrates a real-time reconstruction process compared with M-detector. Our method prevents the accumulation of outdated points from moving objects, leading to much cleaner reconstructions; Fig.~\ref{fig:publicgallery} presents the comparative results with M-detector on the KITTI dataset, emphasizing the advantages of our method in managing crowded traffic scenes in real-world conditions.

\subsection{Application on autonomous scene navigation}
Our system significantly enhances robotic applications: (1) single-frame point cloud reconstruction for collision avoidance and (2) multi-frame fusion of static shapes for real-time scene reconstruction.

As elaborated in Section~\ref{subsec:labeling_scene}, the calculation of free space in a single frame considers both moving and static objects. This ensures that the planned trajectory remains clear of both dynamic and static obstacles. Furthermore, it plays a crucial role in determining whether a given location can be reached via accessible and unobstructed paths, which is vital for effective navigation. The estimated normal vectors for each single-frame point are particularly valuable in identifying ground regions and constraining paths to be suitable for wheel-based robots.

The combined application of our real-time spatial understanding system and robot navigation opens up additional possibilities that merit further exploration. Fig.~\ref{fig:robotnavigation} showcases our system's application for real-time navigation and construction. Fig.~\ref{fig:robotnavigation}(a) compares point-to-point navigation between our method and SLAM based method, i.e., LOAM~\cite{zhang2014loam} and otcomap~\cite{hornung2013octomap}, in real world without priori. Fig.~\ref{fig:robotnavigation}(b) displays synthetic scenes. In Fig.~\ref{fig:robotnavigation}(a) and (b), the first column shows that the robot was stationary for 40 seconds to ensure that the moving object entered the scanned scene, while columns two to five illustrate the process of automatic navigation. The green dot indicates the robot's location and the red dot is the target location. The black dashed curve traces the autonomous exploration path of the robot, while the black solid line indicates the planned path. In our method, free space voxels are color-coded based on the LoS distance field, ranging from low values (purple) to high values (blue). In baseline, the automap is color-coded based on elevation, ranging from low values (purple) to high values (blue). In our system, a robot travels from an initial position to an unknown target location based on real-time free space detection. It utilizes the information on free space to incrementally select waypoints, plan open trajectories to reach the destination, and arrive at the final target without any prior knowledge about the scene. The baseline method stops after encountering a moving object because it treats outdated moving objects as permanent obstacles. 

Fig.~\ref{fig:robotnavigation}(c) shows our system’s application in a virtual coal tunnel exploration task, which is usually under low illuminance. The first and second columns show that the robot handles mobile obstacles, while the third column shows that the robot handles permanent obstacles, even though the obstacle is semantically also a vehicle. The sub-image in the brown frame shows the global surface model of multi-frame reconstruction, and the sub-image in the green frame shows the surface model of single-frame reconstruction. Our method accurately models moving obstacles in real-time single-frame data, as highlighted in the yellow box. In contrast, our method excludes moving obstacles in multi-frame reconstruction and generates a real geographic model. 

Fig.~\ref{fig:robotgallery} presents a gallery showcasing our robot's real-time autonomous exploration and scene reconstruction. Each row depicts the entire process of the robot autonomously exploring and incrementally reconstructing the surface model of the scene. The blue boxes at the bottom of each row offer detailed views of the reconstructed surface in the corresponding local areas. The exploration strategy proposed in~\cite{huang2022autonomous} has been employed, which enables online scanning of previously unknown scenes. Throughout the robot's autonomous navigation and exploration, the surface model of the scanned scene gradually takes shape. Notably, in scenes such as Center and Square, the navigation paths often feature numerous sharp turns due to the presence of obstacles. Our method effectively maintains a smooth and accurate model, highlighting its capacity to handle diverse scenes and to meet the challenges of robotic applications, such as SLAM~\cite{johari2023eslam} and collision avoidance~\cite{lu2023comprehensive}.

\begin{figure}[!t]
\centering
\includegraphics[width=\linewidth]{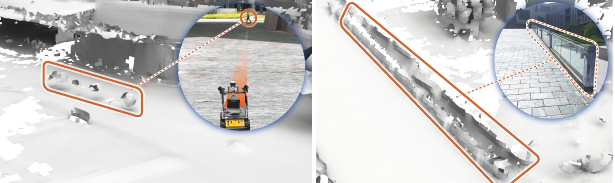}
\caption{
	 Illustration of two failure cases. The red box highlights areas with incorrectly reconstructed surfaces. On the left, the error is due to an insufficiently sampled moving pedestrian from a distance. On the right, the error is caused by a glass wall.}
\label{fig:failurecases}
\end{figure}

\subsection{Limitation and Failure Cases}
Fig.~\ref{fig:failurecases} illustrates two typical failure cases. First, when there are small moving objects in the distance, the 3D point cloud samples captured for them are too sparse for accurate detection, potentially leading to ghost structures. Second, when the scene includes glass surfaces, LiDAR cannot obtain accurate distance readings, resulting in erroneous reconstructions. Reconstruction of these special material objects needs to be supplemented by additional priors.

\section{Conclusion and Future Work}
\label{sec:future}

We propose an intuitive approach to the spatial understanding of real-time LiDAR scan data, simulating the functionality of border cells in the entorhinal-hippocampal circuit. Note that this simulation occurs at the cellular level and thus provides more fundamental information than neural networks, making our method fast and effective. The graphics visibility analysis in implementation has enabled us to completely construct the 3D scene boundaries in single-frame and effectively remove moving objects during the multi-frame reconstruction process. Compared to existing real-time 3D mapping methods such as point-level SDF values, our approach organizes sparse point clouds into continuous regions with spatial properties. This allows the robot to immediately identify safe regions in its environment without having to wait for a complete global map, making it suitable for unknown dynamic scenes.

The evaluation and practical applications of our proposed method have demonstrated its effectiveness. Notably, our system achieves an impressive processing speed of approximately 10Hz, making it well-suited for real-time applications such as robot navigation and autonomous driving. Additionally, our method’s versatility is evident as it does not rely on prior knowledge, assumptions, training data, or high-end GPUs, making it compatible with a wide range of scanning platforms and LiDAR sensor solutions.

In the near future, we will focus on two pivotal areas for further improvement. Firstly, we aim to enhance our moving object tracking capabilities. While our visibility-based moving object removal has proven effective, integrating learning-based completion techniques and data-driven object tracking can enable our system to accurately track and predict the motion of moving objects in real-time. Secondly, we plan to incorporate prior maps into our system. These maps may take various forms, including rough 3D models, elevation maps, or 3D models generated from aerial photography. Leveraging this prior information will allow us to maintain a more compact voxel space and allocate computing resources more efficiently. This approach holds the potential to further enhance reconstruction precision, improve moving object detection, and enable finer reconstruction in complex scenes.

\bibliographystyle{IEEEtran}
\bibliography{RealRecon}

\end{document}